\documentclass[journal]{IEEEtran}

\usepackage[T1]{fontenc}
\usepackage{amsmath,amssymb,amsfonts}
\usepackage{algorithmic}
\usepackage{algorithm}
\usepackage{graphicx}
\usepackage{textcomp}
\usepackage{xcolor}
\usepackage{booktabs}
\usepackage{array}
\usepackage{multirow}
\usepackage{url}
\usepackage{tikz}
\usetikzlibrary{shapes.geometric,arrows.meta,positioning,calc,fit,backgrounds}
\usepackage[hidelinks]{hyperref}
\usepackage[capitalize,noabbrev]{cleveref}

\graphicspath{{figures/}}

\def\BibTeX{{\normalfont B\kern-.05em{\scshape i\kern-.025em b}\kern-.08em
    T\kern-.1667em\lower.7ex\hbox{E}\kern-.125emX}}

\newcommand{\RF}{\textsc{RubricForge}}
\newcommand{\Geval}{\textsc{G-Eval}}
\newcommand{\taubench}{\textsc{$\tau$-bench}}
\newcommand{\webshop}{\textsc{WebShop}}
\newcommand{\rstar}{r^{\star}}

\begin{document}

\title{Inducing Reward-Free Judging Rubrics that Reduce Over-Crediting in Agent Evaluation}

\author{Darragh~Quinn,~David~Dylan,~Roisin~Healy,~Fionn~Carroll,~Maeve~Donnelly,~and~Cormac~Sheehan
\thanks{D.~Dylan, D.~Quinn and M.~Donnelly are with Trinity College Dublin; R.~Healy and C.~Sheehan are with University College Dublin; F.~Carroll is with Dublin City University, Dublin, Ireland.}%
}

\markboth{IEEE Transactions on Knowledge and Data Engineering}%
{Quinn \MakeLowercase{\textit{et al.}}: RubricForge --- Inducing Faithful Evaluation Rubrics for LLM Agents}

\maketitle

\begin{abstract}
Evaluating language-model agents at scale increasingly relies on a second
language model as an automatic judge, because the gold signal, an executable
environment reward, is expensive, slow, or unavailable at deployment time. Such a
judge is a reward-free proxy whose value depends on whether it can be trusted,
yet existing judges either hand-write the scoring rubric, as in \textsc{G-Eval},
or fine-tune the judge's weights, and both tend to credit fluent but unsuccessful
trajectories as successes. We instead induce the text of an agent-judging rubric
from a small set of ground-truth-labeled trajectories, grounding it in true
outcomes. We present \textsc{RubricForge}, which evolves a judge rubric by
reflective evolution against labeled trajectories to maximize agreement with the
environment reward, freezes it, and applies it to held-out trajectories in one
model call with no environment access. The optimized artifact is human-readable
text, so every verdict is attributable to named criteria. Using one frozen 7B model as both agent and judge, on
$\tau$-bench (173 labeled trajectories drawn from 220 rollouts) and
\textsc{WebShop} (160), the principal gain is faithfulness rather than raw
agreement. The edge over a generic \textsc{G-Eval} judge is not statistically
significant (McNemar $p=0.248$), and absolute-score calibration marginally favors
the generic judge ($|{\rm err}|$ difference $-0.048$, $p=2\!\times\!10^{-4}$).
Yet \textsc{RubricForge} over-credits failed trajectories roughly half as often
($0.115$ vs.\ $0.173$ false-pass rate on $\tau$-bench, with three over-credit
catches and zero reversals) and ranks graded \textsc{WebShop} outcomes more
faithfully (Spearman $0.410$ vs.\ $0.370$). For a reward-free evaluator the
false-pass rate, not aggregate agreement, is the deployment-relevant quantity,
since a false pass ships a broken agent whereas a false fail merely costs a
retry.
\end{abstract}


\IEEEpeerreviewmaketitle

\begin{figure*}[t]
  \centering
  \IfFileExists{figures/fig_teaser.pdf}{%
    \includegraphics[width=\textwidth]{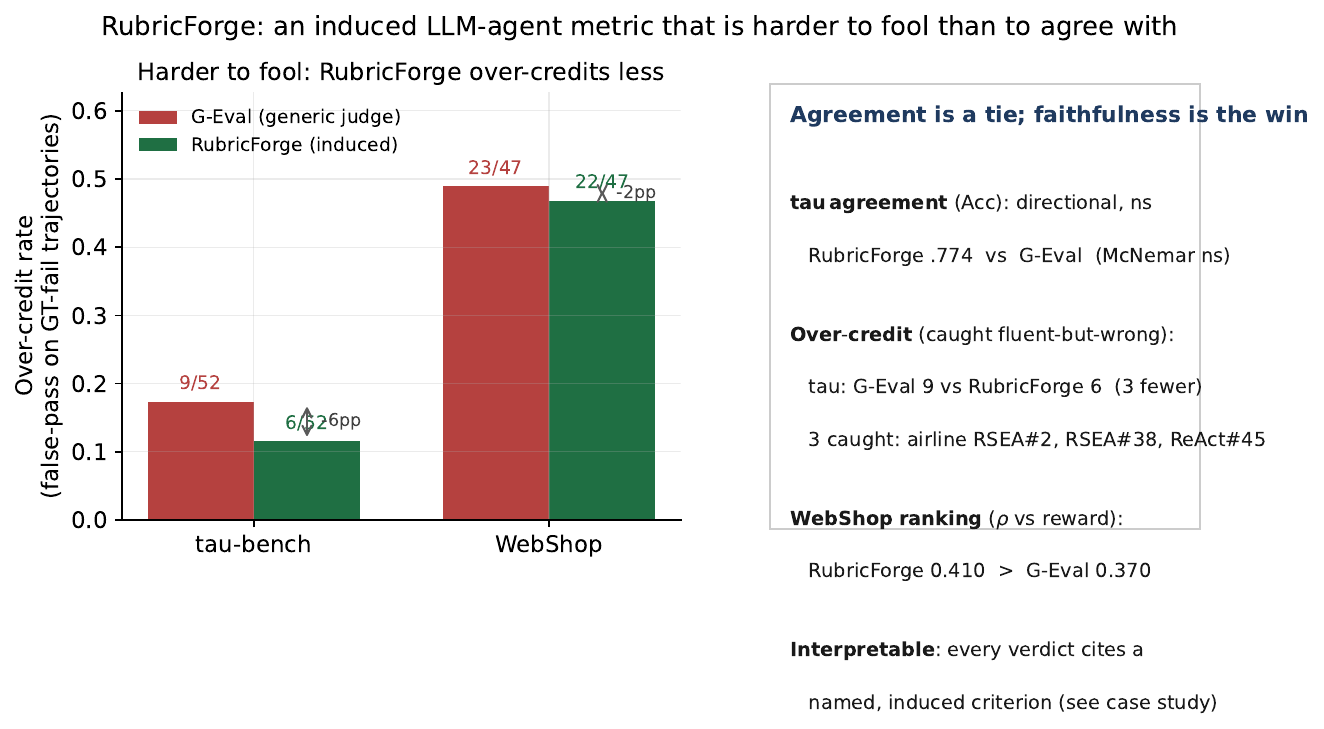}%
  }{%
    \fbox{\parbox[c][4.5cm][c]{0.95\textwidth}{\centering
      \textbf{[ teaser placeholder ]}\\[2pt]
      \texttt{figures/fig\_teaser.pdf} will be inserted here.}}%
  }
  \caption{\RF{} at a glance. An evaluation rubric is \emph{induced} from
  ground-truth-labeled agent trajectories by reflective evolution, then frozen
  and applied to held-out trajectories with one model call and no environment
  access. Left: against a generic \Geval{} judge, the binary-agreement edge is
  directional and \emph{not} statistically significant
  (McNemar $p=0.248$). Right: where it matters, the induced rubric over-credits
  fluent-but-failed trajectories about half as often
  ($0.115$ vs.\ $0.173$ false-pass rate on \taubench{}), the deployment-relevant
  error for a reward-free evaluator. \RF{} is thus \emph{harder to fool than to
  agree}.}
  \label{fig:teaser}
\end{figure*}

\section{Introduction}\label{sec:intro}
\IEEEPARstart{L}{anguage}-model agents---systems that interleave reasoning with
tool calls to act in an external environment---are now evaluated, debugged, and
ranked at a scale that no longer admits human grading of every
trajectory~\cite{yao2023react,shinn2023reflexion,wang2023voyager,su2025autonomysurvey}.
The cleanest evaluation signal is an \emph{environment reward}: a programmatic
oracle that inspects the world state after an episode and returns success or a
graded score---a database hash in a customer-service simulator, an attribute
match in a shopping environment, a passing test suite for a code
agent~\cite{yao2024taubench,yao2022webshop,jimenez2024swebench}. Such oracles
are the ground truth this paper is built on. But they are also the very thing
one rarely has when it is most needed. During development one wants to score
thousands of candidate rollouts cheaply; in deployment the environment is the
production system itself, where a destructive ``ground-truth'' probe (issuing
the refund, placing the order, mutating the record) is exactly what must
\emph{not} be run merely to grade the agent. The field has therefore turned to
the \emph{LLM-as-a-judge}: a second language model that reads a trajectory and
predicts whether the agent
succeeded~\cite{zheng2023judging,liu2023geval,gu2024judgesurvey}. The judge is a
\emph{reward-free proxy}---a learned stand-in for an oracle it never queries.

A proxy is only as useful as it is trustworthy, and judges are known to fail in
characteristic ways: they prefer verbose, fluent, confident
answers~\cite{wang2024notfair}, they favor their own
generations~\cite{panickssery2024selfpref}, and they reward surface form over
substance. For an \emph{agent} judge these biases concentrate into one
particularly costly mistake. A trajectory can read beautifully---the agent
authenticates the user, narrates a plan, and signs off with a polished
confirmation---while the underlying action failed: the booking errored out, the
record was never committed, the purchased item does not match the request. A
judge that rewards the narration \emph{over-credits} the failure. We argue that
this single error type, not aggregate agreement, is what determines whether a
reward-free evaluator is safe to deploy, because the two ways a binary judge can
be wrong are not symmetric. A \emph{false pass}---scoring a truly-failed
trajectory as success---silently certifies and ships a broken agent: it inflates
reported success rates, masks regressions, and selects bad policies during
optimization. A \emph{false fail}---scoring a truly-successful trajectory as
failure---merely costs a retry or a discarded sample. The risk is asymmetric, so
the metric should be too.

This paper studies whether the \emph{text} of an agent-judging rubric should be
\emph{written} or \emph{induced}. The dominant practice writes the rubric by
hand---\Geval{} prompts a strong model with human-authored
criteria~\cite{liu2023geval}---or fine-tunes the judge's
\emph{weights}~\cite{kim2024prometheus,kim2024prometheus2,zhu2025judgelm}. We
take a third path. We keep the judge model \emph{frozen} and instead optimize
the rubric \emph{string} by reflective evolution against a small set of
ground-truth-labeled trajectories, using each trajectory's true environment
outcome as the supervision signal. The induced rubric is then frozen and applied
to a held-out, by-task-disjoint split with a single model call and no
environment access. We call this \RF{}. Because the artifact under optimization
is human-readable text, the resulting metric is interpretable by construction:
every verdict is attributable to named criteria the practitioner can inspect,
edit, and audit. And because induction grounds the rubric in true outcomes
rather than in the judge's untethered intuitions, we hypothesize that it is more
resistant to over-crediting fluent failures.

Our central result separates two questions that the literature usually
conflates: whether the judge agrees with the oracle in aggregate, and whether it
fails in the dangerous direction. On a hard binary success label the frozen 7B
judge is near its agreement ceiling, and \RF{} does not significantly out-agree a
generic \Geval{} judge; on graded \webshop{} the generic judge is in fact
marginally better calibrated in absolute score. We report both outcomes without
qualification. On the false-pass rate, however, \RF{} over-credits fluent
failures roughly half as often, and the entire margin between the two judges on
\taubench{} reduces to three trajectories, all of which are over-credit catches
in \RF{}'s favor with zero reversals. Aggregate agreement is therefore the wrong
headline for a reward-free evaluator: induction leaves it essentially unchanged
while sharply reducing the error that determines whether the proxy is safe to
deploy.

This result speaks to a broader question about optimizing in the space of
prompts. When the objective being optimized is decoupled from real outcomes, an
optimized prompt can drift or collapse; grounding the same optimization in
checkable signal is what keeps it useful. \RF{} grounds rubric induction in
labeled trajectory outcomes, which is precisely what lets an induced proxy resist
drifting away from the true reward, and we develop this connection to Goodhart's
law in \cref{sec:related} and \cref{sec:discussion}.

\medskip
\noindent\textbf{Contributions.} The main contributions of this paper are
summarized as follows.
\begin{itemize}
  \item \textbf{\RF{}: automatic metric induction for agents.} We introduce a
  method that evolves a judge rubric against ground-truth-labeled trajectories
  with a \emph{frozen} backbone, then freezes the rubric and scores held-out
  trajectories with one model call and no environment access
  (\cref{sec:method}). Induction reliably improves validation agreement
  ($0.654\!\to\!0.769$ on \taubench{}, $0.667\!\to\!0.750$ on \webshop{}) and
  produces concrete, human-readable, trajectory-grounded criteria.
  \item \textbf{Faithfulness, not agreement, is where induction pays off.} We
  show \RF{} over-credits failed trajectories about half as often as a generic
  judge ($0.173\!\to\!0.115$ false-pass on \taubench{}; three fluent failures
  caught, zero reversed), never scoring worse than \Geval{} on any item---while
  stating explicitly that the binary-agreement margin is directional and
  \emph{not} significant (McNemar $p=0.248$).
  \item \textbf{Better ranking of graded outcomes.} \RF{} ranks graded
  \webshop{} outcomes more faithfully (Spearman $0.410$ vs.\ $0.370$), even
  though absolute calibration marginally favors the generic judge ($|{\rm err}|$
  difference $-0.048$, $p=2\!\times\!10^{-4}$). We separate ranking faithfulness
  from absolute calibration and report both.
  \item \textbf{A reward-free evaluation protocol and an over-crediting probe.}
  We package an oracle-reward ceiling, a false-pass rate, a
  ranking-versus-calibration split, per-criterion leave-one-out attribution, and
  difficulty/fluency/length stratification into a reusable protocol
  (\cref{sec:analysis}), and connect the false-pass rate to proxy-objective
  misalignment and Goodhart's law.
\end{itemize}

\noindent\textbf{Roadmap.} \Cref{sec:related} surveys LLM-as-a-judge, reward
modeling, agent benchmarks, prompt optimization, and proxy misalignment.
\Cref{sec:problem} formalizes reward-free trajectory evaluation.
\Cref{sec:method} presents \RF{} with an algorithm and an architecture diagram.
\Cref{sec:setup} details datasets, baselines, and protocol; \cref{sec:results}
reports the main table; \cref{sec:analysis} is the faithfulness diagnosis.
\Cref{sec:discussion} discusses implications and limitations, and
\cref{sec:conclusion} concludes.

\section{Related Work}\label{sec:related}
We organize the literature into five themes: (a) LLM-as-a-judge and automatic
evaluation, where the false-pass failure our probe measures is named; (b) reward
modeling, RLHF/RLAIF, and learned evaluators; (c) agent benchmarks and
evaluation; (d) prompt optimization and evolution; and (e) proxy/metric
misalignment and Goodhart's law. Throughout, we mark the differentiator of
\RF{}: prior judges either \emph{hand-write} the rubric or \emph{fine-tune} the
judge's weights, whereas \RF{} \emph{induces} the rubric \emph{text} by
evolution while keeping the judge model frozen.

\subsection{LLM-as-a-Judge and Automatic Evaluation}\label{sec:rw:judge}
Using a strong language model to grade the outputs of another has become the
default scalable alternative to human evaluation. \Geval{} prompts a capable
model with a hand-written rubric and a chain-of-thought form-filling protocol to
score natural-language generation~\cite{liu2023geval}; MT-Bench and Chatbot
Arena established LLM and human pairwise judging as a benchmark substrate for
chat models~\cite{zheng2023judging}. A complementary line trains dedicated
\emph{open} evaluators: Prometheus and Prometheus~2 fine-tune models to apply
fine-grained, user-supplied score rubrics and to approximate
GPT-4-level judgments~\cite{kim2024prometheus,kim2024prometheus2}; JudgeLM,
PandaLM, and Auto-J fine-tune scalable judges for instruction-following and
pairwise comparison~\cite{zhu2025judgelm,wang2024pandalm,li2024autoj};
FLASK decomposes evaluation into skill sets~\cite{ye2024flask}; and aggregation
schemes such as branch-solve-merge and panels of diverse juror models reduce
single-judge variance~\cite{saha2024branchsolvemerge,verga2024replacing}. The
same machinery is increasingly turned on \emph{agents} rather than single
responses---PersonaGym scores persona-conditioned agents and CharacterEval grades
multi-turn role-play behavior~\cite{samuel2024personagym,tu2024charactereval}---
which is the regime \RF{} targets. Surveys catalog the rapidly growing
space~\cite{gu2024judgesurvey,li2024fromgeneration,li2024llmsasjudges}.

Crucially for us, the same line documents that judges are \emph{biased
estimators} that fail in a specific, dangerous direction. LLM evaluators are not
fair: they are swayed by answer position, verbosity, and superficial
fluency~\cite{wang2024notfair}; they recognize and favor their own
generations~\cite{panickssery2024selfpref}. These biases are exactly what
produce a \emph{false pass}---scoring a fluent-but-failed trajectory as a
success---which is the error our over-crediting probe isolates and measures.
Viewed through this lens, an induced agent rubric is a \emph{learned proxy} for
the true environment reward, and the false-pass rate is the operational
measurement of the gap between that proxy and the
reward~\cite{yang2026misalignment}. \RF{} differs from all of the above in
\emph{what} is optimized. \Geval{} fixes a human-written rubric; Prometheus and
JudgeLM move the judge's weights. \RF{} leaves the judge model frozen and
optimizes the rubric \emph{string} against ground-truth labels---cheaper than
fine-tuning, more faithful than hand-writing, and interpretable by construction
because the artifact is text.

\subsection{Reward Modeling, RLHF, and Learned Evaluators}\label{sec:rw:reward}
Learned evaluators have a long lineage in preference-based alignment. Reward
models trained on human comparisons drive RLHF for summarization and instruction
following~\cite{stiennon2020summarize,ouyang2022instructgpt}; RLAIF replaces or
augments the human signal with AI feedback and constitutional
self-critique~\cite{lee2024rlaif,bai2022constitutional}; and direct preference
optimization recasts the language model itself as an implicit reward
model~\cite{rafailov2023dpo}. ``Self-rewarding'' and ``meta-rewarding'' schemes
let a model generate its own training rewards or judge its own
judgments~\cite{yuan2024selfrewarding,wu2024metarewarding}, and self-feedback
loops such as Self-Refine and tool-interactive critique iteratively improve an
output against the model's own criticism~\cite{madaan2023selfrefine,gou2023critic}.
The reliability of
these learned rewards is itself now benchmarked, e.g.\ RewardBench measures how
well reward models track held-out
preferences~\cite{lambert2024rewardbench}. \RF{} shares the goal of a learned
evaluator but targets a different object and supervision. Reward models score
\emph{responses} from \emph{scalar preference} data and are typically consumed
by an optimizer; \RF{} scores entire \emph{agent trajectories} against
\emph{programmatic outcome} labels and is consumed as a frozen, reward-free
\emph{measurement} instrument whose verdicts are human-auditable rather than a
black-box scalar.

\subsection{Agent Benchmarks and Evaluation}\label{sec:rw:agents}
The agents we judge come from interactive benchmarks with executable rewards.
$\tau$-bench scores tool-agent-user dialogues in retail and airline
customer-service domains by hashing the final database state against a gold
state, yielding a strict binary reward~\cite{yao2024taubench}. \webshop{}
provides a graded reward measuring how well a purchased product matches a natural
-language instruction's attributes and price~\cite{yao2022webshop}. Broader
suites---AgentBench across eight environments~\cite{liu2024agentbench}, WebArena
for realistic web tasks~\cite{zhou2024webarena}, GAIA for general
assistants~\cite{mialon2023gaia}, and SWE-bench for repository-level
coding~\cite{jimenez2024swebench}---all rely on programmatic success checks. The
trajectories themselves are produced by agent policies such as
ReAct~\cite{yao2023react}, reflective and tree-structured
variants~\cite{shinn2023reflexion,yao2023tot}, experiential and
memory-augmented agents~\cite{zhao2024expel,wang2024awm,packer2023memgpt},
generative agents~\cite{park2023generative}, reasoning bootstrappers~\cite{zelikman2022star},
and self-evolving or recursively-composed
agents~\cite{wang2023voyager,su2025autonomysurvey,yang2026recursivemas}. Many of
these agents are moreover \emph{retrieval-augmented}, grounding their actions in
fetched evidence rather than parametric memory
alone~\cite{lewis2020rag,karpukhin2020dpr,asai2024selfrag,shi2024replug}, which
only widens the space of fluent-but-unfaithful traces a reward-free judge must
screen. These benchmarks supply the
ground-truth labels \RF{} induces against; \RF{} is orthogonal to them, learning
a reward-free proxy that predicts their oracle from the trajectory alone, so it
can grade rollouts when re-running the oracle is too costly or unsafe.

\subsection{Prompt Optimization and Evolution}\label{sec:rw:promptopt}
\RF{} induces its rubric with reflective prompt evolution rather than gradient
descent. Automatic prompt search spans Monte-Carlo instruction
generation (APE)~\cite{zhou2023ape}, optimization-by-prompting where an LLM
proposes improved prompts from a trajectory of past
attempts (OPRO)~\cite{yang2024opro}, evolutionary prompt
search (PromptBreeder)~\cite{fernando2023promptbreeder}, and the DSPy line that
compiles and optimizes language-model
pipelines~\cite{khattab2022dsp,khattab2024dspy}. We specifically build on
GEPA, which uses natural-language \emph{reflection} over execution traces to
mutate prompts and has been shown to rival reinforcement learning at far lower
sample cost~\cite{agrawal2025gepa}. \RF{} applies this machinery to a target it
was not previously used for: the single mutated component is the \emph{judge's
own rubric}, the ``rollout'' is a \emph{judging pass} rather than an environment
rollout, and the fitness is \emph{agreement with the ground-truth label}. The
optimizer is reused unchanged; the contribution is the reward-free induction
\emph{objective}, not a new search algorithm.

Reflective prompt evolution sits within a broader family of \emph{self-improvement
without weight updates}. Models can bootstrap from their own
high-confidence generations~\cite{huang2022selfimprove}, marginalize over sampled
reasoning paths~\cite{wang2023selfconsistency}, and revise prior attempts---though
the limits of unaided self-correction are now well documented: it helps with
external feedback but can fail or even hurt when the model grades
itself~\cite{huang2024selfcorrect,kamoi2024selfcorrection}, which is precisely why
\RF{} grounds its reflection in \emph{external} ground-truth labels rather than the
model's own confidence. The induction stage is also a form of \emph{test-time}
adaptation: it spends extra compute to fit an instrument before deployment,
echoing test-time scaling~\cite{snell2024testtime,muennighoff2025s1} and the
classical meta-learning and test-time-adaptation
program~\cite{finn2017maml,nichol2018reptile,hospedales2021survey,min2022metaicl,sun2020ttt,wang2021tent}
of learning-to-adapt from limited supervision---here adapting a \emph{rubric},
not weights.

\subsection{Proxy/Metric Misalignment and Goodhart's Law}\label{sec:rw:goodhart}
That an optimized proxy drifts from the true objective is Goodhart's law: once a
measure becomes a target it ceases to be a good
measure~\cite{goodhart1984problems}. In machine learning this manifests as
reward over-optimization and reward hacking---policies that exploit an imperfect
reward to score highly while violating the
intent~\cite{amodei2016concrete,skalse2022defining,pan2022effects}, with
characterized scaling laws for how proxy reward and true reward
diverge under optimization pressure~\cite{gao2023scaling}. A reward-free judge is
itself a proxy at risk of this drift: an ungrounded judge that learns to reward
fluency \emph{is} a Goodharted metric. Our over-crediting probe is a direct,
operational measurement of this drift for evaluation metrics---the false-pass
rate is how far the proxy has slipped toward rewarding form over outcome. The
enabling contrast is grounding. ToolTree grounds an agent's planning in
per-instance, programmatically checkable
feedback~\cite{yang2026tooltree}; analogously, \RF{} grounds rubric
\emph{induction} in per-trajectory ground-truth labels, which is exactly why the
induced metric resists drifting toward fluency where an ungrounded generic judge
silently does. Grounding a small-sample objective in real outcomes rather than an
internal preference is the same move that few-shot preference alignment
makes~\cite{zhao2023gpo}. This grounded stance contrasts with ungrounded prompt-space optimization, in which
an aggregate objective decoupled from real outcomes can drift or collapse; here a
grounded objective instead confers faithfulness.

\section{Problem Formulation}\label{sec:problem}
We formalize reward-free trajectory evaluation and state the induction
objective. \Cref{tab:notation} summarizes the notation.

\begin{table}[t]
\centering
\caption{Notation used throughout the paper.}
\label{tab:notation}
\footnotesize
\renewcommand{\arraystretch}{1.18}
\begin{tabular}{@{}l p{0.66\columnwidth}@{}}
\toprule
Symbol & Meaning \\
\midrule
$\tau$ & An agent trajectory (actions, observations, final answer) \\
$x(\tau)$ & Judge-readable text rendering of $\tau$ \\
$\rstar(\tau)$ & Environment (ground-truth) reward: binary or graded \\
$y(\tau)$ & Binary success label, $y=\mathbf{1}[\rstar>0]$ \\
$\mathcal{D}$ & Pool of labeled trajectories $\{(\tau_i, \rstar_i)\}$ \\
$\mathcal{D}_{\rm tr},\mathcal{D}_{\rm va},\mathcal{D}_{\rm te}$ & Train / val / test splits (by-task disjoint) \\
$g_\theta$ & Frozen judge backbone with parameters $\theta$ \\
$\rho$ & Rubric string (the judge's system prompt) \\
$M_\rho$ & Induced metric: $M_\rho(\tau)\!=\!(\hat p,\hat s,\hat z)$ \\
$\hat p\in\{0,1\}$ & Predicted pass verdict \\
$\hat s\in[0,1]$ & Predicted graded score \\
$\hat z$ & Free-text rationale (one sentence) \\
$\mathcal{C}=\{c_1,\dots,c_K\}$ & Induced criteria parsed from $\rho$ \\
$A(\rho;\mathcal{D})$ & Agreement of $M_\rho$ with labels on $\mathcal{D}$ \\
$\mathrm{FP}(\rho)$ & False-pass rate on truly-failed trajectories \\
\bottomrule
\end{tabular}
\end{table}

\noindent\textbf{Setup.} An agent acting in an environment produces a trajectory
$\tau$---a sequence of actions (tool calls and user-facing messages) interleaved
with observations (tool results and user replies), terminating in a final
committed answer. After the episode an oracle returns the environment reward
$\rstar(\tau)$, either binary (e.g.\ a database-state hash in
$\tau$-bench~\cite{yao2024taubench}) or graded in $[0,1]$ (e.g.\ the attribute
-match score in \webshop{}~\cite{yao2022webshop}). We define the binary success
label $y(\tau)=\mathbf{1}[\rstar(\tau)>0]$. The oracle is the ground truth; the
defining constraint of \emph{reward-free} evaluation is that at scoring time the
evaluator may \emph{not} query it, because doing so is expensive, slow, or, in a
live system, destructive.

\noindent\textbf{The induced metric.} A reward-free evaluator is a function that
maps the trajectory \emph{text} $x(\tau)$ to a prediction without environment
access. We realize it as a frozen judge backbone $g_\theta$ conditioned on a
rubric string $\rho$:
\begin{equation}
\label{eq:metric}
M_\rho(\tau) \;=\; g_\theta\big(\rho,\, x(\tau)\big) \;=\; (\hat p,\, \hat s,\, \hat z),
\end{equation}
a single deterministic call (temperature $0$) returning a pass verdict
$\hat p\in\{0,1\}$, a graded score $\hat s\in[0,1]$, and a one-sentence rationale
$\hat z$. The text rendering $x(\tau)$ exposes only what an observer of the
interaction would see---actions, tool results, user replies, counters, and the
final answer---and never the oracle reward or gold world-state, so no label
leaks into the judge's input.

\noindent\textbf{Objective.} Let agreement on a labeled set be
\begin{equation}
\label{eq:agreement}
A(\rho;\mathcal{D}) \;=\; \frac{1}{|\mathcal{D}|}\sum_{\tau\in\mathcal{D}}
\mathbf{1}\!\big[\hat p_\rho(\tau) = y(\tau)\big].
\end{equation}
Rubric induction seeks the rubric maximizing agreement on the labeled training
pool, with selection on a held-out validation pool to avoid overfitting the
exact training trajectories:
\begin{equation}
\label{eq:induction}
\rho^{\star} \;=\; \arg\max_{\rho\,\in\,\mathcal{R}}\; A\big(\rho;\, \mathcal{D}_{\rm tr}\big),
\quad \text{selected by } A\big(\rho;\,\mathcal{D}_{\rm va}\big),
\end{equation}
over the space $\mathcal{R}$ of candidate rubric strings reachable by the
evolution operator. The judge backbone $\theta$ is \emph{frozen}: optimization
moves only the text $\rho$, not the weights. The frozen $\rho^{\star}$ is then
applied once per held-out trajectory in $\mathcal{D}_{\rm te}$.

\noindent\textbf{What we actually optimize for.} Agreement (\cref{eq:agreement})
is the \emph{training} signal, but it is not the deployment-relevant quantity. We
separate two error directions. On a truly-failed trajectory ($y=0$), a
\emph{false pass} ($\hat p = 1$) silently certifies a broken agent; on a
truly-successful trajectory ($y=1$), a \emph{false fail} merely costs a retry.
We therefore single out the false-pass rate,
\begin{equation}
\label{eq:fp}
\mathrm{FP}(\rho) \;=\; \frac{\big|\{\tau:\, y(\tau)=0 \ \wedge\ \hat p_\rho(\tau)=1\}\big|}
{\big|\{\tau:\, y(\tau)=0\}\big|},
\end{equation}
as the primary \emph{faithfulness} criterion, and treat raw agreement,
$\kappa$, AUC, ranking (Spearman), and absolute calibration as complementary
diagnostics. The central empirical question is then sharp: does grounding the
rubric in labels (\cref{eq:induction}) reduce $\mathrm{FP}$
(\cref{eq:fp})---make the proxy harder to fool---even where it does not move raw
agreement?

\section{Method: RubricForge}\label{sec:method}
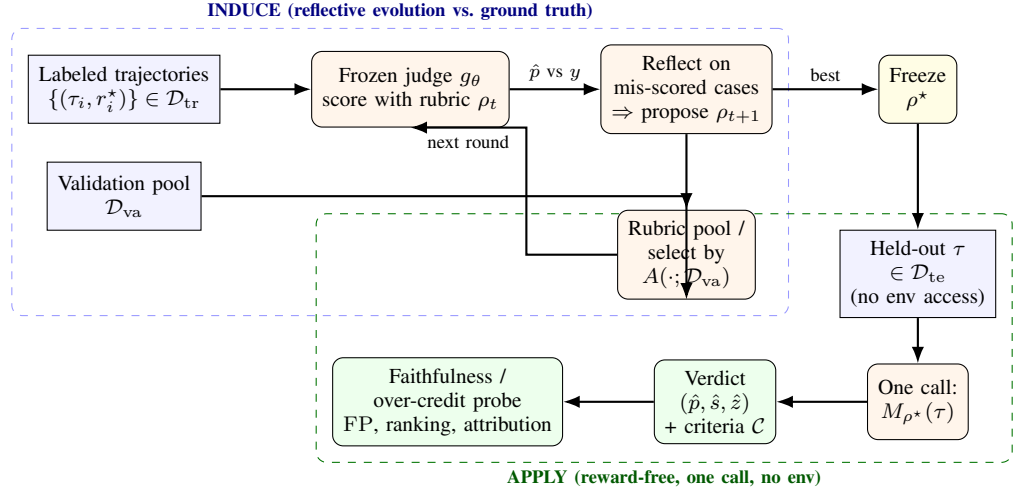
\begin{figure*}[t]
\centering
\footnotesize
\begin{tikzpicture}[
  node distance=8mm and 12mm,
  box/.style={draw, rounded corners, align=center, inner sep=4pt, minimum height=9mm, font=\footnotesize},
  data/.style={draw, align=center, inner sep=4pt, minimum height=9mm, font=\footnotesize, fill=blue!5},
  proc/.style={draw, rounded corners, align=center, inner sep=4pt, minimum height=10mm, font=\footnotesize, fill=orange!8},
  outp/.style={draw, rounded corners, align=center, inner sep=4pt, minimum height=9mm, font=\footnotesize, fill=green!7},
  arr/.style={-{Latex[length=2.2mm]}, thick},
  every node/.style={font=\footnotesize}
]
\node[data] (traj) {Labeled trajectories\\$\{(\tau_i,\rstar_i)\}\in\mathcal{D}_{\rm tr}$};
\node[data, below=5mm of traj] (val) {Validation pool\\$\mathcal{D}_{\rm va}$};
\node[proc, right=of traj] (judge1) {Frozen judge $g_\theta$\\score with rubric $\rho_t$};
\node[proc, right=of judge1] (reflect) {Reflect on\\mis-scored cases\\$\Rightarrow$ propose $\rho_{t+1}$};
\node[box, below=10mm of reflect, fill=orange!8] (pool) {Rubric pool /\\select by\\$A(\cdot;\mathcal{D}_{\rm va})$};
\node[box, right=14mm of reflect, fill=yellow!12] (freeze) {Freeze\\$\rho^{\star}$};
\node[data, below=14mm of freeze] (test) {Held-out $\tau$\\$\in\mathcal{D}_{\rm te}$\\(no env access)};
\node[proc, below=6mm of test] (apply) {One call:\\$M_{\rho^{\star}}(\tau)$};
\node[outp, left=12mm of apply] (verdict) {Verdict\\$(\hat p,\hat s,\hat z)$\\+ criteria $\mathcal{C}$};
\node[outp, left=12mm of verdict] (probe) {Faithfulness /\\over-credit probe\\$\mathrm{FP}$, ranking, attribution};

\draw[arr] (traj) -- (judge1);
\draw[arr] (judge1) -- node[above, font=\scriptsize]{$\hat p$ vs $y$} (reflect);
\draw[arr] (reflect) -- (pool);
\draw[arr] (pool.west) -- ($(pool.west)+(-1.2,0)$) |- (judge1.south)
  node[pos=0.75, below, font=\scriptsize]{next round};
\draw[arr] (val.east) -| (pool.south);
\draw[arr] (reflect) -- node[above, font=\scriptsize]{best} (freeze);
\draw[arr] (freeze) -- (test);
\draw[arr] (test) -- (apply);
\draw[arr] (apply) -- (verdict);
\draw[arr] (verdict) -- (probe);

\begin{scope}[on background layer]
  \node[draw=blue!40, dashed, rounded corners, fit=(traj)(val)(judge1)(reflect)(pool),
        inner sep=6pt, label={[blue!50!black,font=\scriptsize\bfseries]north:INDUCE (reflective evolution vs.\ ground truth)}] {};
  \node[draw=green!45!black, dashed, rounded corners, fit=(test)(apply)(verdict)(probe),
        inner sep=6pt, label={[green!35!black,font=\scriptsize\bfseries]south:APPLY (reward-free, one call, no env)}] {};
\end{scope}
\end{tikzpicture}
\caption{\RF{} architecture. \textbf{Induce} (top, blue): the frozen judge
$g_\theta$ scores training trajectories with the current rubric $\rho_t$; a
reflection step reads the cases whose verdict disagreed with the ground-truth
label and proposes an improved rubric $\rho_{t+1}$; candidates are pooled and
selected by validation agreement (\cref{eq:induction}). \textbf{Freeze}: the
best rubric $\rho^{\star}$ is frozen. \textbf{Apply} (bottom, green): for each
held-out trajectory $M_{\rho^{\star}}$ emits a verdict $(\hat p,\hat s,\hat z)$
in a single call with no environment access, feeding the faithfulness /
over-crediting probe. The judge backbone is never updated; only the rubric text
is optimized.}
\label{fig:arch}
\end{figure*}

\RF{} has three stages: \emph{induce} a rubric by reflective evolution against
labeled trajectories, \emph{freeze} it, and \emph{apply} it to held-out
trajectories as a one-call reward-free metric. \Cref{fig:arch} diagrams the
pipeline and \cref{alg:rubricforge} states it.

\subsection{Stage 1: Rubric Induction}
The optimizer is GEPA's reflective prompt
evolution~\cite{agrawal2025gepa}, reused unchanged; \RF{}'s novelty is in the
component, the rollout, and the fitness it is pointed at. The single evolved
component is the rubric string $\rho$ (the judge's system prompt). The
``rollout'' for a candidate rubric is not an environment rollout but a
\emph{judging pass}: the frozen backbone scores each training trajectory with
$\rho$, and the per-trajectory fitness is binary agreement with the gold label,
\begin{equation}
\label{eq:roll}
s(\tau;\rho) \;=\; \mathbf{1}\!\big[\hat p_\rho(\tau) = y(\tau)\big],
\end{equation}
so the optimizer's objective is exactly agreement (\cref{eq:agreement}).

The reflection step is where grounding enters. After a minibatch is scored, the
mis-scored trajectories---those where the judge's verdict disagreed with the true
label---are collected, and the reflection meta-prompt presents the current
rubric together with these failures and their true labels, asking for a revised
rubric that would label them correctly and generalize. Two design choices keep
the induction honest and leakage-free. First, the rubric author (the reflecting
model) is told the true label \emph{only for the training trajectories it is
revising}, and is explicitly instructed to describe \emph{observable trajectory
evidence} (which signals indicate the user's needs were met versus the agent
guessing) rather than to mention the reward---so the evolved rubric encodes
\emph{symptoms of success}, not a memorized answer key, and the deployed judge
never sees a label. Second, the seed of the search is not an empty or strawman
rubric but the same competent generic G-Eval-style rubric used by our \Geval{}
baseline; any rubric the search accepts must therefore beat a real judge, not a
degenerate one, so reported gains are not an artifact of a weak starting point.
Candidates are pooled and the rubric maximizing validation agreement
(\cref{eq:induction}) is retained.

\subsection{Stage 2: Freeze}
The selected rubric $\rho^{\star}$ is frozen verbatim. If the search never beats
its seed, \RF{} explicitly records that the effective rubric is the generic
default; in our runs the search did improve over the seed on both benchmarks
(\cref{sec:results}), accepting evolved rubrics into pools of size $6$
($\tau$-bench) and $4$ (\webshop{}). Freezing makes the metric a fixed,
reproducible instrument: the same trajectory always receives the same verdict.

\subsection{Stage 3: Reward-Free Application}
At test time the frozen $\rho^{\star}$ is the judge's system prompt and the
rendered held-out trajectory $x(\tau)$ is the user message; a single
deterministic call returns the verdict (\cref{eq:metric}). A fixed output
contract, kept separate from the evolved rubric so the optimizer shapes criteria
rather than formatting, requests a strict JSON object
$\{\text{pass},\text{score},\text{reason}\}$. Parsing is robust---the first
brace-delimited object is extracted and decoded---and \emph{conservative}: any
network or parse failure falls back to $\hat p = 0$, so a malformed judge call
never spuriously \emph{passes} a trajectory (it can only abstain toward failure,
the safe direction for a reward-free evaluator). Because the rubric is text, its
named criteria $\mathcal{C}$ can be applied individually to attribute each
verdict to the criterion responsible, which we exploit in \cref{sec:analysis}.

\begin{algorithm}[t]
\caption{\RF{}: reward-free metric induction}
\label{alg:rubricforge}
\begin{algorithmic}[1]
\REQUIRE labeled pools $\mathcal{D}_{\rm tr},\mathcal{D}_{\rm va}$; frozen judge $g_\theta$; seed rubric $\rho_0$; budget $T$; minibatch $b$
\STATE \textbf{// Stage 1: induce}
\STATE pool $\gets \{\rho_0\}$;\; evaluate $A(\rho_0;\mathcal{D}_{\rm va})$
\FOR{$t = 1$ \TO $T$}
  \STATE sample parent $\rho$ from pool; draw minibatch $B\subset\mathcal{D}_{\rm tr}$, $|B|=b$
  \STATE for each $\tau\in B$: $(\hat p,\hat s,\hat z)\gets g_\theta(\rho, x(\tau))$;\; $s(\tau)\gets\mathbf{1}[\hat p=y(\tau)]$
  \STATE $F \gets \{\tau\in B : \hat p \neq y(\tau)\}$ \COMMENT{mis-scored cases}
  \STATE $\rho' \gets \textsc{Reflect}(\rho,\, F,\, \text{true labels of } F)$ \COMMENT{describe symptoms, not the label}
  \IF{$A(\rho';\mathcal{D}_{\rm va}) \geq \max_{\rho\in\text{pool}} A(\rho;\mathcal{D}_{\rm va})$}
     \STATE pool $\gets$ pool $\cup\,\{\rho'\}$
  \ENDIF
\ENDFOR
\STATE \textbf{// Stage 2: freeze}
\STATE $\rho^{\star} \gets \arg\max_{\rho\in\text{pool}} A(\rho;\mathcal{D}_{\rm va})$
\STATE \textbf{// Stage 3: apply (reward-free, no env access)}
\STATE \textbf{return} $M_{\rho^{\star}}: \tau \mapsto g_\theta(\rho^{\star}, x(\tau))$
\end{algorithmic}
\end{algorithm}

\section{Experimental Setup}\label{sec:setup}
\noindent\textbf{Benchmarks and trajectories.} We evaluate on two agent
benchmarks with executable rewards. \taubench{}~\cite{yao2024taubench} covers
retail and airline customer service with a strict \emph{binary}
database-state-hash reward; we dump $220$ labeled rollouts and admit them
\emph{whole tasks at a time} (all of a task's cells stay in one split, so no task
leaks across the train/val/test boundary) into a class-stratified
$85/26/62$ split, which fills its budget with the $173$ cells reported below; the
remaining $47$ surplus cells are held out. The $173$ enter the experiment at a
$\sim$$15\%$ positive rate. \webshop{}~\cite{yao2022webshop} provides a
\emph{graded} attribute-match reward; we collect $160$ trajectories, of which
$116$ carry graded scores. For each benchmark we dump trajectories from two agent
policies---the standard ReAct loop~\cite{yao2023react} and a self-evolving RSEA
policy---so the judge sees both terse and elaborated traces. Each trajectory is
rendered to judge-readable text exposing only observable signals (actions, tool
results and errors, user replies, step/error counters, and the final committed
answer); the oracle reward and gold world-state are withheld from the renderer.

\noindent\textbf{Frozen backbone.} A single frozen Qwen2.5-7B model serves as
\emph{both} the agent backbone and the judge, at temperature $0$ for
deterministic, reproducible verdicts. Using one frozen 7B model for everything is
a deliberate stress test: it removes any capability gap between agent and judge
and isolates the effect of the \emph{rubric} rather than a stronger grader.

\noindent\textbf{Splits.} Trajectories are partitioned \emph{by task} into
train/validation/test at roughly $50/15/35$, so no test task appears in
induction. On \taubench{} the $173$ admitted cells yield $85$ train
($12$ positive), $26$ validation ($4$ positive), and $62$ test trajectories
($10$ positive, $\sim16\%$), summing to $85\!+\!26\!+\!62=173$ at an overall
$\sim15\%$ positive rate; \webshop{} uses $80$ train and $24$ validation, with
$56$ graded test trajectories for the calibration analysis.
\Cref{tab:datastats} consolidates these per-split trajectory counts, positive
rates, reward types, and induction budgets in one place; note that the test
positive rate ($\sim$$16\%$) is held close to the pool rate ($\sim$$15\%$) by the
by-task class-stratified split, so the agreement numbers below are not an artifact
of a skewed test label.

\begin{table}[t]
\centering
\caption{Dataset and protocol statistics. Trajectories are split \emph{by task}
(no task crosses a split boundary); positive rate is $\Pr[\rstar>0]$. ``Graded''
counts the subset carrying a continuous $\rstar\in[0,1]$ used for the
calibration analysis. Budget is GEPA judging calls during induction.}
\label{tab:datastats}
\footnotesize
\renewcommand{\arraystretch}{1.18}
\setlength{\tabcolsep}{4.5pt}
\begin{tabular}{@{}l c c@{}}
\toprule
 & \taubench{} & \webshop{} \\
\midrule
Reward type & binary (db-hash) & graded $[0,1]$ \\
Rollouts dumped & $220$ & $160$ \\
Admitted into split & $173$ & $160$ \\
\midrule
Train (\#\,/\,\#pos) & $85$\,/\,$12$ & $80$\,/\,--- \\
Validation (\#\,/\,\#pos) & $26$\,/\,$4$ & $24$\,/\,--- \\
Test (\#\,/\,\#pos) & $62$\,/\,$10$ & $56$ graded \\
Test positive rate & $\sim$$16\%$ & graded \\
Overall positive rate & $\sim$$15\%$ & --- \\
\midrule
Agent policies & ReAct\,/\,RSEA & ReAct\,/\,RSEA \\
Induction budget (calls) & $200$ ($208$ used) & $150$ ($156$ used) \\
Accepted rubric pool & $6$ & $4$ \\
Best val.\ agreement & $0.769$ ($20/26$) & $0.750$ ($18/24$) \\
\bottomrule
\end{tabular}
\end{table}

\noindent\textbf{Baselines.} We compare \RF{} against (i) \emph{Oracle}, the
environment reward itself (a ceiling, agreement $1.0$ by construction); (ii)
\Geval{}~\cite{liu2023geval}, the generic hand-written judge---the same rubric
that seeds induction, so the comparison isolates the value of grounding; (iii)
\emph{FewShot}, the generic rubric augmented with $k\!=\!4$ in-context labeled
trajectory examples (no rubric evolution); (iv) \emph{Heuristic}, a non-LLM
surface-feature rule; and (v) \emph{Majority}, the train-majority class predicted
for every item (degenerate but a standard sanity floor on an imbalanced label).

\noindent\textbf{Metrics.} For binary agreement we report Accuracy with a
bootstrap $95\%$ confidence interval, $F_1$ on the positive class, Cohen's
$\kappa$ (which, unlike accuracy, penalizes the degenerate all-negative
predictor), and ROC-AUC. For graded outcomes we report Spearman's $\rho$ and
Kendall's $\tau$ (ranking faithfulness) and mean absolute error $|\hat s -
\rstar|$ (absolute calibration). The faithfulness primary metric is the
false-pass rate $\mathrm{FP}$ (\cref{eq:fp}).

\noindent\textbf{Significance.} Every judge-vs-judge comparison carries a paired
test: McNemar's exact test on per-item agreement for binary verdicts, and a
paired bootstrap on absolute error for graded scores. We mark significance as
\texttt{***}~$p\!<\!0.001$, \texttt{**}~$p\!<\!0.01$, \texttt{*}~$p\!<\!0.05$,
and \texttt{ns} otherwise, and we label directional-but-not-significant gaps
\emph{ns} explicitly rather than implying they are wins. Confidence intervals are
$10^4$-resample nonparametric bootstrap.

\section{Results}\label{sec:results}
\Cref{tab:main} reports the main \taubench{} test results. We read it through
finding-titled observations rather than a single aggregate verdict, because the
honest story is split across metrics.

\begin{table*}[t]
\centering
\caption{Main results on \taubench{} test ($n=62$, $\sim$$16\%$ positive). Judge
$\leftrightarrow$ ground-truth agreement. Best non-degenerate value per column in
\textbf{bold}. Majority (all-fail) and Heuristic (pass-almost-everything, acc.\
$0.161$) are degenerate predictors; their inflated metrics are bracketed and
excluded from the per-column best. McNemar column tests each judge against \RF{} on per-item agreement
($p$, significance). $\uparrow$: higher is better.}
\label{tab:main}
\footnotesize
\renewcommand{\arraystretch}{1.22}
\begin{tabular}{@{}lccccc@{}}
\toprule
Judge & Acc.\ $\uparrow$ [95\% CI] & $F_1$ $\uparrow$ & Cohen $\kappa$ $\uparrow$ & AUC $\uparrow$ & McNemar vs.\ \RF{} \\
\midrule
Oracle (env.\ reward) & $1.000$ [$1.000,1.000$] & $1.000$ & $1.000$ & $1.000$ & --- \\
\midrule
\textbf{\RF{} (ours)} & $\mathbf{0.774}$ [$0.661,0.871$] & $\mathbf{0.222}$ & $\mathbf{+0.092}$ & $0.485$ & --- \\
\Geval{} (generic) & $0.726$ [$0.613,0.839$] & $0.190$ & $+0.026$ & $0.500$ & $p=0.248$~\texttt{ns} \\
FewShot ($k=4$) & $0.613$ [$0.484,0.726$] & $0.200$ & $-0.019$ & $\mathbf{0.535}$ & $p=0.009$~\texttt{**} \\
Heuristic & $0.161$ [$0.081,0.258$] & [$0.278$] & $\phantom{+}0.000$ & $0.500$ & $p<0.001$~\texttt{***} \\
Majority (degenerate) & [$0.839$] [$0.742,0.919$] & $0.000$ & $\phantom{+}0.000$ & $0.500$ & $p=0.289$~\texttt{ns} \\
\bottomrule
\end{tabular}
\end{table*}

\noindent\textbf{Finding 1: Induction works, and the rubric is concrete.}
Reflective evolution improves validation agreement from the generic seed in both
domains: $0.654\!\to\!0.769$ on \taubench{} (a pool of $6$ rubrics, $208$
judging calls) and $0.667\!\to\!0.750$ on \webshop{} (a pool of $4$, $156$
calls). The induced \taubench{} rubric is not an opaque tuned string but five
named, trajectory-grounded criteria---successful authentication (no repeated
failed attempts), order verification (no ``order not found''), request
fulfillment with explicit user confirmation, error-free tool usage, and clear
user communication---each phrased in terms of observable trajectory evidence
(\cref{app:rubric}).

\noindent\textbf{Finding 2: The binary-agreement edge is real in direction but
not significant.} \RF{} attains the highest accuracy among informative judges
($0.774$ vs.\ $0.726$ for \Geval{}) and the highest informative Cohen
$\kappa$ ($+0.092$ vs.\ $+0.026$); $\kappa$ is the honest headline here because
it discounts the trivial benefit of predicting the majority class on a $16\%$
-positive label. But the McNemar test against \Geval{} is \emph{not} significant
($p=0.248$): on a hard binary label, a frozen 7B judge is near its agreement
ceiling, and we do not claim a significant binary win. We say so in the abstract
and here.

\noindent\textbf{Finding 3: Majority is degenerate; $\kappa$ exposes it.} The
Majority baseline posts the highest \emph{accuracy} in the table ($0.839$) purely
by predicting ``fail'' for every trajectory on an imbalanced label---yet its
$\kappa$ and $F_1$ are both $0$, and McNemar versus \RF{} is not significant
($p=0.289$). This is the canonical reason accuracy alone is the wrong headline
metric for reward-free evaluation: a metric that never identifies a success is
useless for ranking agents, which $\kappa$ correctly reflects and accuracy hides.

\noindent\textbf{Finding 4: FewShot and Heuristic are clearly worse.} Adding
four in-context examples without rubric evolution \emph{hurts} ($0.613$ accuracy,
$\kappa=-0.019$; McNemar $p=0.009$, \texttt{**}): the demonstrations bias the
judge toward over-passing fluent traces (its predicted-positive rate jumps to
$0.32$). The non-LLM Heuristic collapses to near-zero accuracy
($p<0.001$, \texttt{***}). Neither is a competitive reward-free evaluator. These
two significant gaps confirm the test has power; the non-significant \Geval{} gap
is therefore a genuine near-tie on binary agreement, not low power---which is
exactly why the next section moves the analysis to where induction does
separate.

\noindent\textbf{Finding 5: The error decomposition shows the tie is one-sided.}
\Cref{tab:fullmetrics} unpacks each informative judge's $62$ test verdicts into
the full confusion count (TP/FP/FN/TN) together with its predicted-positive rate
$\hat{p}_{+}$, false-pass count, and false-fail count. The decomposition makes the
``near-tie'' precise and reveals it is \emph{asymmetric}. \RF{} and \Geval{} agree
on their true-positive count ($2$) and---critically---on their false-fail count
($8$ each): induction does \emph{not} cost a single extra retry. The whole
difference is in false passes, which fall from $9$ (\Geval{}) to $6$ (\RF{}),
dragging the predicted-positive rate from $0.177$ toward the gold $\sim$$0.16$
($0.129$ for \RF{}). FewShot moves the opposite way---its demonstrations push
$\hat{p}_{+}$ up to $0.323$ and its false passes to $17$, the mechanism behind its
significant accuracy loss. Reading the table by \emph{error direction} rather than
by aggregate accuracy is what exposes that the \Geval{} tie is entirely a
false-pass tie that \RF{} wins; \cref{fig:confusion} visualizes the same
decomposition.

\begin{table*}[t]
\centering
\caption{Error-direction decomposition on \taubench{} test ($n=62$; $10$ pass,
$52$ fail). Each row splits the judge's verdicts into true/false
positives/negatives, its predicted-positive rate $\hat{p}_{+}$, the
deployment-relevant false-pass count/rate (\cref{eq:fp}), and the merely-costly
false-fail count. \RF{} and \Geval{} share an identical false-fail count ($8$); the
entire margin is the $3$ fewer false passes, so \RF{} is \emph{never worse on any
item}. Degenerate predictors (Heuristic, Majority) are bracketed. Best informative
value per column in \textbf{bold}; for $\hat{p}_{+}$ ``best'' is closest to the
gold positive rate $0.161$.}
\label{tab:fullmetrics}
\footnotesize
\renewcommand{\arraystretch}{1.2}
\setlength{\tabcolsep}{6pt}
\begin{tabular}{@{}l ccccc cc c c@{}}
\toprule
Judge & TP & FP & FN & TN & $\hat{p}_{+}$ & Acc.\ $\uparrow$ & Cohen $\kappa$ $\uparrow$ & \#\,false-pass $\downarrow$ & FP rate $\downarrow$ \\
\midrule
Oracle (env.\ reward) & $10$ & $0$ & $0$ & $52$ & $0.161$ & $1.000$ & $1.000$ & $0$ & $0.000$ \\
\midrule
\textbf{\RF{} (ours)} & $2$ & $\mathbf{6}$ & $8$ & $\mathbf{46}$ & $\mathbf{0.129}$ & $\mathbf{0.774}$ & $\mathbf{+0.092}$ & $\mathbf{6}$ & $\mathbf{0.115}$ \\
\Geval{} (generic) & $2$ & $9$ & $8$ & $43$ & $0.177$ & $0.726$ & $+0.026$ & $9$ & $0.173$ \\
FewShot ($k=4$) & $3$ & $17$ & $7$ & $35$ & $0.323$ & $0.613$ & $-0.019$ & $17$ & $0.327$ \\
Heuristic & [$10$] & [$52$] & [$0$] & [$0$] & [$1.000$] & $0.161$ & $\phantom{+}0.000$ & [$52$] & [$1.000$] \\
Majority (degenerate) & [$0$] & [$0$] & [$10$] & [$52$] & [$0.000$] & [$0.839$] & $\phantom{+}0.000$ & [$0$] & [$0.000$] \\
\bottomrule
\end{tabular}
\end{table*}

\section{Analysis: Faithfulness and Over-Crediting}\label{sec:analysis}
If binary agreement does not separate \RF{} from a generic judge, what does? This
section answers with the faithfulness probe: the over-crediting (false-pass) rate
and its case study, per-criterion attribution, the ranking-versus-calibration
split, and stratification by difficulty, fluency, and length.

\subsection{The Over-Crediting Probe}
\begin{figure}[t]
\centering
\IfFileExists{figures/fig_overcredit.pdf}{%
  \includegraphics[width=\columnwidth]{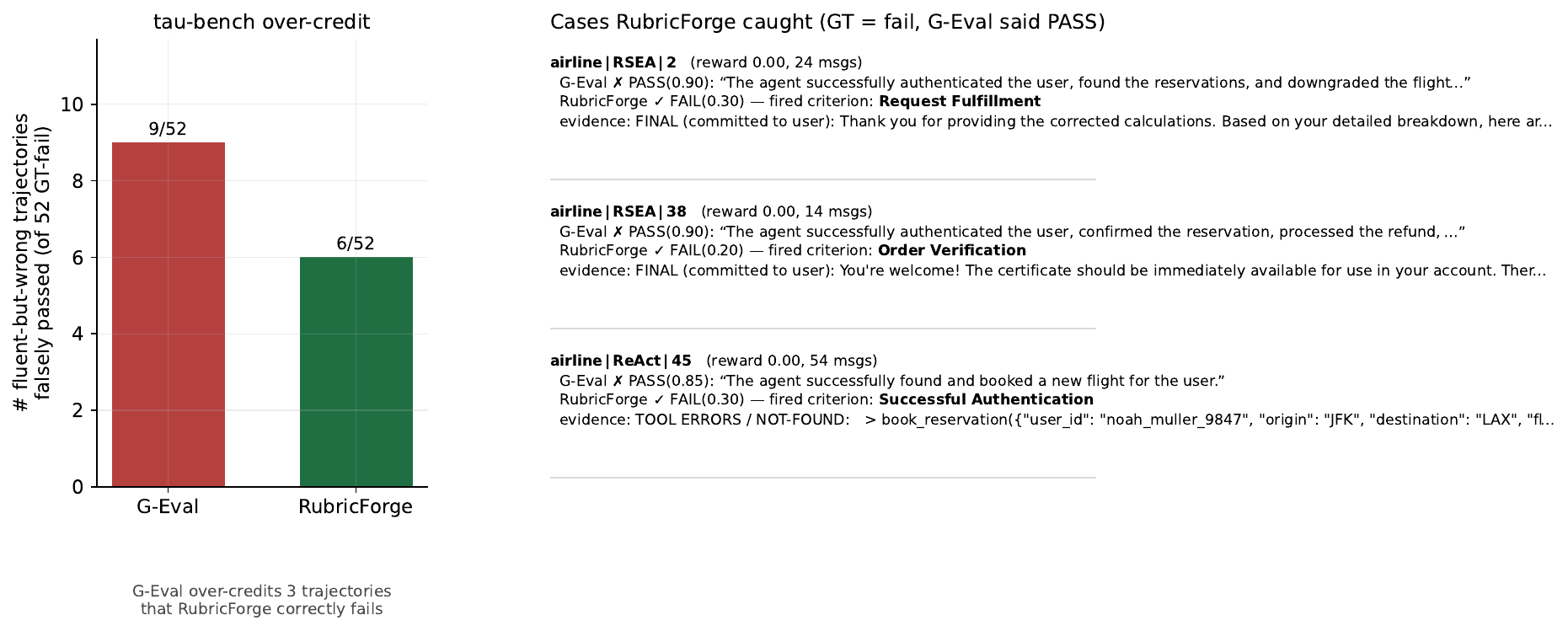}}{%
  \fbox{\parbox[c][3.2cm][c]{0.92\columnwidth}{\centering\texttt{fig\_overcredit.pdf}}}}
\caption{Over-crediting (false-pass) rate on truly-failed trajectories. \RF{}
over-credits fluent-but-failed trajectories about half as often as \Geval{} on
\taubench{} ($0.115$ vs.\ $0.173$) and slightly less often on \webshop{}
($0.468$ vs.\ $0.489$). On \taubench{} the only three judge disagreements are
three \Geval{} false-passes that \RF{} catches, with zero reversals.}
\label{fig:overcredit}
\end{figure}

\Cref{fig:overcredit} reports the false-pass rate (\cref{eq:fp}): how often each
judge scores a \emph{truly-failed} trajectory as a success. On \taubench{}, of
$52$ ground-truth failures \RF{} over-credits $6$ ($\mathrm{FP}=0.115$) versus
\Geval{}'s $9$ ($\mathrm{FP}=0.173$)---about half as often. Decisively, the two
judges disagree on \emph{exactly three} \taubench{} items, and all three are
\Geval{} false-passes that \RF{} correctly fails, with \emph{zero} reversals
(no item where \RF{} over-credits and \Geval{} does not). The entire margin
between the two judges in \cref{tab:main} \emph{is} these three over-credit
catches: \RF{} is never worse than \Geval{} on any individual item. On graded
\webshop{} the gap is narrower but in the same direction: of $47$ failures \RF{}
over-credits $22$ ($0.468$) versus \Geval{}'s $23$ ($0.489$), with $2$ caught and
$1$ reversed. The takeaway is the paper's thesis in one number: induction's
benefit is concentrated in the dangerous error direction even where it is
invisible in aggregate agreement.

\begin{figure}[t]
\centering
\IfFileExists{figures/fig_confusion.pdf}{%
  \includegraphics[width=\columnwidth]{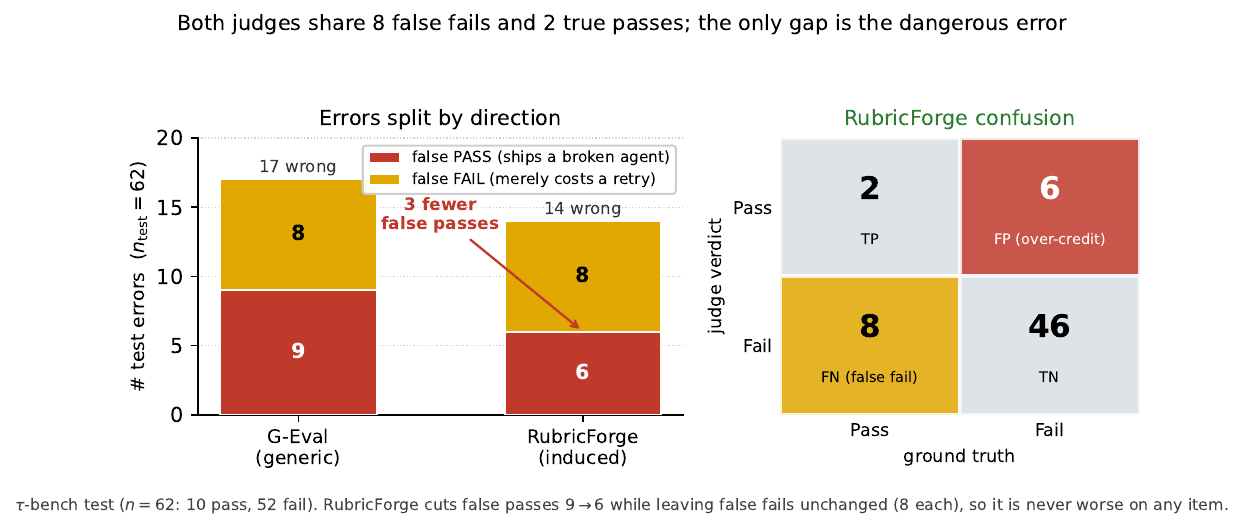}}{%
  \fbox{\parbox[c][3.2cm][c]{0.92\columnwidth}{\centering\texttt{fig\_confusion.pdf}}}}
\caption{Error-direction decomposition on \taubench{} test ($n=62$). Left:
verdicts stacked by error type---both judges make the same $8$ false fails
(amber), so the only difference is the dangerous false-pass bar (red), which
\RF{} cuts $9\!\to\!6$. Right: \RF{}'s confusion grid; the $3$ caught cases leave
the over-credit (FP) cell for the true-fail (TN) cell while the false-fail (FN)
cell is unchanged. \RF{} is thus never worse than \Geval{} on any item.}
\label{fig:confusion}
\end{figure}

\Cref{fig:confusion} recasts the same result as a two-way error decomposition.
Stacking each judge's $62$ test verdicts by error \emph{direction} (left) shows
the false-fail bars are identical height ($8$ each) while the false-pass bar
shrinks from $9$ to $6$; the \RF{} confusion grid (right) makes explicit that the
$3$ caught cases all move out of the over-credit (FP) cell and into the true-fail
(TN) cell, with the false-fail (FN) cell untouched. This is the visual statement
of ``harder to fool than to agree'': the only cell that changes is the dangerous
one, and it changes in the safe direction.

\subsection{Case Study: The Three Caught Trajectories}
Because the \taubench{} margin reduces to three trajectories, we can inspect them
exhaustively (\cref{tab:cases}). Each is a fluent airline trajectory that
\Geval{} passes with high confidence ($\hat s \in [0.85, 0.90]$) and \RF{}
fails---and in each case the induced rubric pins the failure to a specific named
criterion. In \texttt{airline$|$RSEA$|$2} the agent narrates a savings breakdown
from downgrades but never commits the update; \RF{}'s \emph{Request Fulfillment}
criterion fires (``did not confirm \dots\ or ensure the downgrades were
completed''). In \texttt{airline$|$RSEA$|$38} the agent assures the user a
certificate is ``immediately available'' though the order was never verified;
\emph{Order Verification} fires. In \texttt{airline$|$ReAct$|$45} the
\texttt{book\_reservation} call returns ``payment amount does not add up,'' yet
the agent reports ``successfully booked''; \emph{Successful Authentication} (and,
in the breakdown, Order Verification and Request Fulfillment) fire. In all three,
\Geval{} rewards the confident final message; \RF{} reads the trajectory evidence
the rubric tells it to look for. This is the interpretability dividend of
inducing \emph{text}: the metric does not merely disagree, it says \emph{which
criterion} the trajectory violated.

\begin{table}[t]
\centering
\caption{The only three \taubench{} judge disagreements (all $\rstar=$ fail). In
every case \Geval{}{=}\textsc{pass} (over-credit) and \RF{}{=}\textsc{fail}
(correct), with the induced criterion that fired.}
\label{tab:cases}
\scriptsize
\renewcommand{\arraystretch}{1.25}
\begin{tabular}{@{}p{0.28\columnwidth}p{0.40\columnwidth}p{0.18\columnwidth}@{}}
\toprule
Trajectory & Why it truly failed & Criterion fired \\
\midrule
\texttt{airline$|$RSEA$|$2} & Narrates a savings breakdown from downgrades but never commits the update & Request Fulfillment \\
\texttt{airline$|$RSEA$|$38} & Claims certificate ``immediately available''; order never verified & Order Verification \\
\texttt{airline$|$ReAct$|$45} & Booking call returns ``payment amount does not add up'' yet the agent reports ``booked'' & Successful Authentication \\
\bottomrule
\end{tabular}
\end{table}

\subsection{Per-Criterion Leave-One-Out Ablation}
\begin{figure}[t]
\centering
\IfFileExists{figures/fig_ablation.pdf}{%
  \includegraphics[width=\columnwidth]{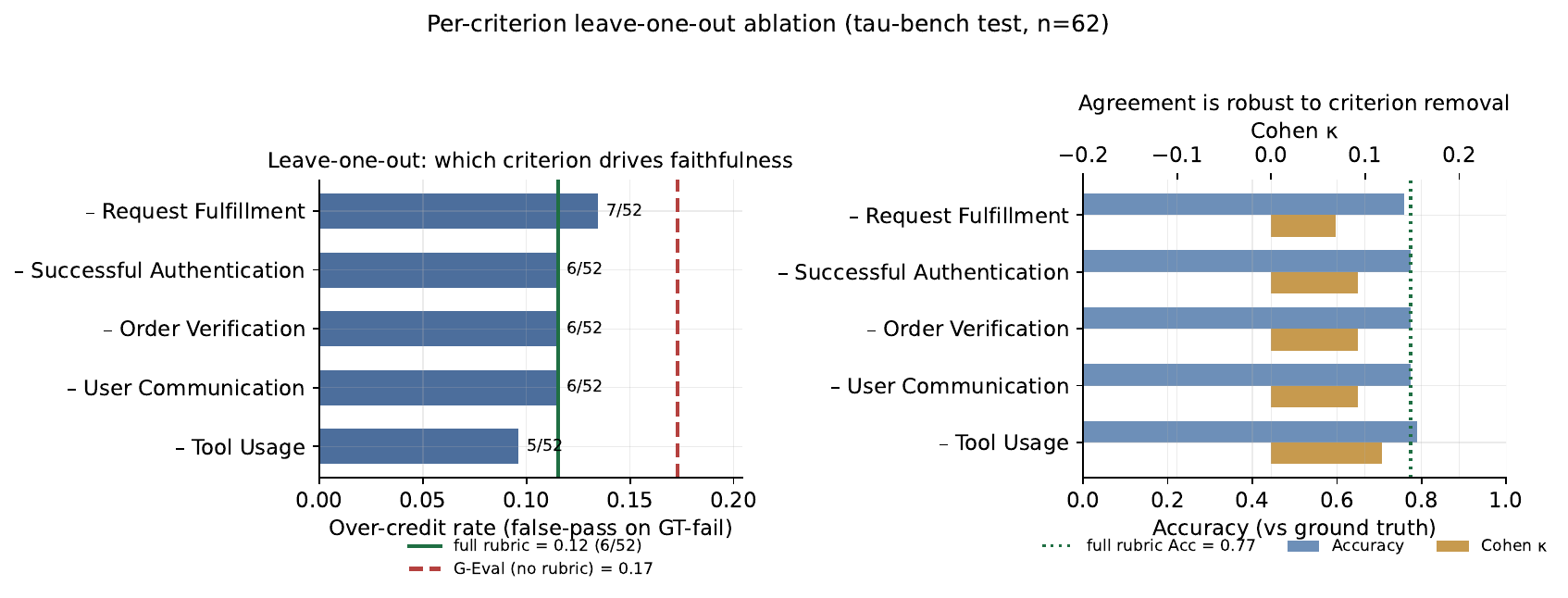}}{%
  \fbox{\parbox[c][3.2cm][c]{0.92\columnwidth}{\centering\texttt{fig\_ablation.pdf}}}}
\caption{Per-criterion leave-one-out on \taubench{} test. Removing \emph{Request
Fulfillment} worsens every faithfulness metric (false-pass $0.115\!\to\!0.135$,
$\kappa$ $+0.092\!\to\!+0.068$): it most drives faithfulness. Removing \emph{Tool
Usage} is mildly redundant; the remaining criteria leave the test metrics
unchanged.}
\label{fig:ablation}
\end{figure}

Which criteria carry the faithfulness? We re-run the frozen judge with each of
the five induced criteria removed in turn and re-measure accuracy, $\kappa$, and
the false-pass rate on \taubench{} test (\cref{fig:ablation}). Removing
\emph{Request Fulfillment} is the only ablation that clearly \emph{degrades}
faithfulness: accuracy falls $0.774\!\to\!0.758$, $\kappa$ falls
$+0.092\!\to\!+0.068$, and the false-pass rate \emph{worsens}
$0.115\!\to\!0.135$---this criterion, which demands explicit user confirmation
that the requested action was completed, is what most resists over-crediting.
Removing \emph{Tool Usage} is mildly redundant: accuracy nudges to $0.790$,
$\kappa$ to $+0.118$, and false-pass to $0.096$ (the tool-error signal it
captures is partly subsumed by the other criteria). Removing Authentication,
Order Verification, or User Communication leaves accuracy, $\kappa$, and
false-pass exactly unchanged on this test set. The rubric is thus robust to
single-criterion removal, with a clear faithfulness driver and one mildly
redundant criterion---an interpretability statement no opaque judge can make.

\subsection{Ranking versus Calibration}
\begin{figure}[t]
\centering
\IfFileExists{figures/fig_calibration.pdf}{%
  \includegraphics[width=\columnwidth]{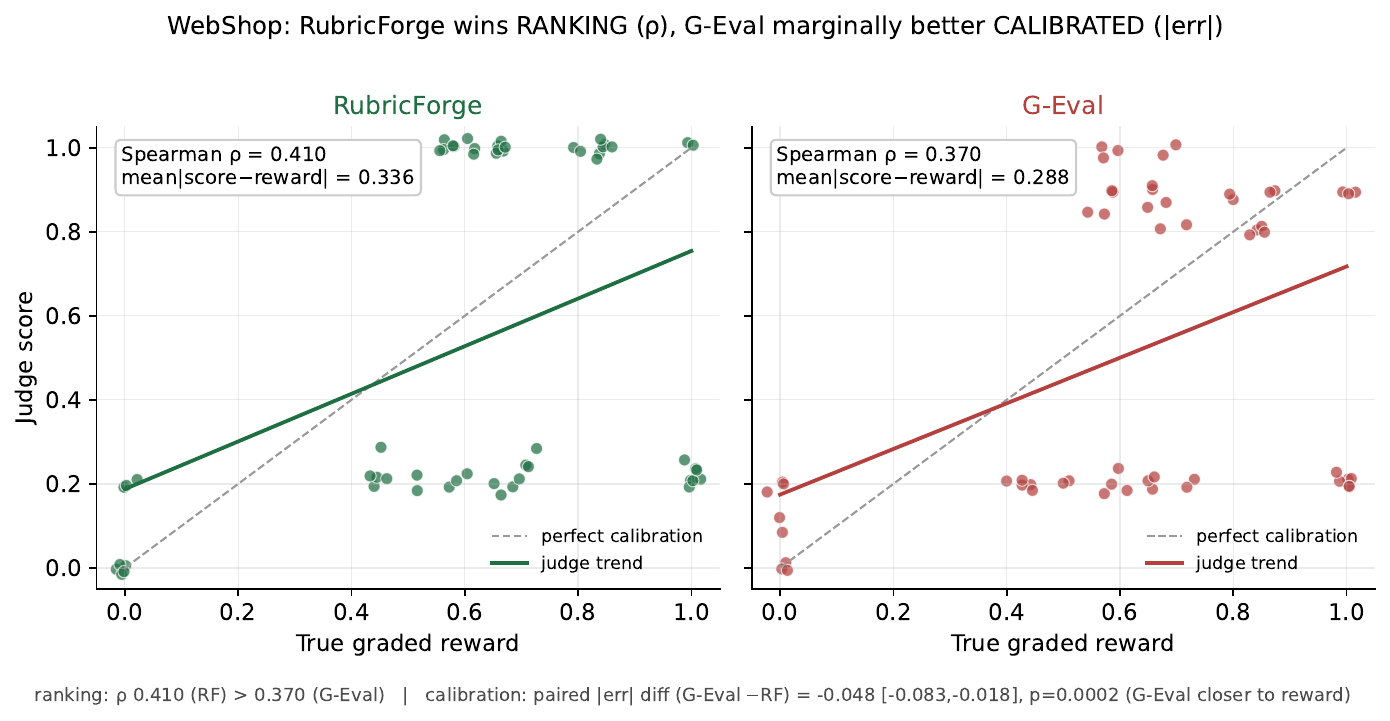}}{%
  \fbox{\parbox[c][3.2cm][c]{0.92\columnwidth}{\centering\texttt{fig\_calibration.pdf}}}}
\caption{Ranking versus calibration on graded \webshop{} ($n=56$). \RF{} ranks
graded outcomes better (Spearman $0.410$ vs.\ $0.370$) but \Geval{} is marginally
better calibrated in absolute error (mean $|\hat s-\rstar|$ $0.288$ vs.\ $0.336$;
paired-bootstrap $\Delta=-0.048$, $p=2\!\times\!10^{-4}$). Ranking faithfulness
and absolute calibration are distinct, and we report both honestly.}
\label{fig:calibration}
\end{figure}

On graded \webshop{} we separate two often-conflated notions of score quality:
\emph{ranking} (does a higher judge score imply a higher true reward?) and
\emph{absolute calibration} (is the score numerically close to the reward?).
\Cref{fig:calibration} shows they come apart, and honesty requires reporting both
directions. \RF{} \emph{wins ranking}: Spearman $0.410$ (Kendall $0.347$) versus
\Geval{}'s $0.370$ (Kendall $0.314$)---its scores order graded outcomes more
faithfully, which is what one needs to compare or select agents. But \Geval{} is
marginally better \emph{calibrated} in absolute error: mean $|\hat s-\rstar|$ is
$0.288$ for \Geval{} versus $0.336$ for \RF{}, and a paired bootstrap on absolute
error favors \Geval{} by $0.048$ ($95\%$ CI $[0.018, 0.083]$,
$p=2\!\times\!10^{-4}$). We do not paper over this: \RF{}'s evolved rubric tends
toward more decisive $\{0.2, 1.0\}$-style scores, which sharpens ranking but
inflates absolute error against the smooth graded reward. The two metrics measure
different things, and a reward-free evaluator used for \emph{agent comparison}
should be judged on ranking---where \RF{} wins---while one used as an absolute
score regressor should not.

\subsection{Stratification: Where the Advantage Lives}
Finally we ask \emph{where} \RF{}'s over-crediting advantage concentrates, since
a uniform gap and a localized one have different implications. Stratifying the
\taubench{} false-pass gap (\cref{tab:strata}) shows the advantage is not
uniform. It concentrates in the \emph{harder airline} domain (false-pass gap
$+0.19$: \Geval{} $0.438$ vs.\ \RF{} $0.250$), and within it in the
\emph{fluent} self-evolving \texttt{airline$|$RSEA} policy (gap $+0.25$:
\Geval{} $0.375$ vs.\ \RF{} $0.125$), whose elaborated, confident narration is
exactly what fools a generic judge. The easier retail domain shows no gap (both
$0.056$): there, fluent-failure over-crediting is rare for either judge, so there
is little to win. Critically, the advantage is \emph{not} a length artifact: in
the longest-trajectory tertile the false-pass gap is $0.00$ (both judges
over-credit nothing there), while the gap is largest in shorter, surface-fluent
traces. \RF{}'s faithfulness edge is therefore driven by surface-fluency
$\times$ task-difficulty---precisely the regime where a reward-free evaluator is
most likely to be fooled---rather than by trajectory length.

\begin{table}[t]
\centering
\caption{\taubench{} false-pass rate stratified. The over-crediting advantage
concentrates in the hard \emph{airline} domain and the fluent \texttt{RSEA}
policy, \emph{not} in the longest trajectories. Gap $=$ \Geval{} $-$ \RF{}
(higher $=$ larger \RF{} advantage).}
\label{tab:strata}
\footnotesize
\renewcommand{\arraystretch}{1.2}
\setlength{\tabcolsep}{4pt}
\begin{tabular}{@{}p{0.30\columnwidth}cccc@{}}
\toprule
Stratum & $n_{\rm fail}$ & \Geval{} FP & \RF{} FP & Gap \\
\midrule
Retail & $36$ & $0.056$ & $0.056$ & $\phantom{+}0.00$ \\
Airline (harder) & $16$ & $0.438$ & $0.250$ & $+0.19$ \\
\texttt{airline$|$RSEA} (fluent) & $8$ & $0.375$ & $0.125$ & $+0.25$ \\
\midrule
Length T1 ($\le$$30$, short) & $18$ & $0.278$ & $0.167$ & $+0.11$ \\
Length T3 ($>$$54$, longest) & $17$ & $0.000$ & $0.000$ & $\phantom{+}0.00$ \\
\bottomrule
\end{tabular}
\end{table}

\section{Discussion}\label{sec:discussion}
\noindent\textbf{The false-pass rate is the deployment-relevant metric.} The
through-line of our results is that for a reward-free evaluator, aggregate
agreement is the wrong headline and the false-pass rate is the right one. A
metric that ships broken agents (high false-pass) is dangerous regardless of its
accuracy on an imbalanced label; a metric that occasionally discards a good
sample (false-fail) is merely inefficient. \RF{} does not significantly win the
former-irrelevant binary-agreement contest, and we say so plainly---but it nearly
halves the false-pass rate, the quantity that actually governs whether one can
trust a reported success number. ``Harder to fool than to agree'' is not a slogan
covering a weak result; it is the precise, honest description of \emph{which}
axis induction improves.

\noindent\textbf{Grounding the objective resists Goodhart.} Why does inducing the
rubric against labels help where it matters? Because it grounds the metric's
objective in true outcomes. An ungrounded judge optimizes an internal notion of
``looks successful,'' which a fluent agent can satisfy without succeeding---a
Goodharted metric drifting toward
form~\cite{goodhart1984problems,gao2023scaling}. \RF{}'s reflection step is
repeatedly shown trajectories that \emph{looked} successful but were labeled
failures, and is pushed to articulate the observable evidence that distinguishes
them---which is exactly an anti-over-crediting pressure. This is the
\emph{grounded dual} of the sibling negative result (\cref{sec:rw:goodhart}):
where an ungrounded prompt-space meta-objective \emph{collapses} across users, a
grounded evaluation objective confers faithfulness, and just as grounding an
agent's planning in checkable feedback sharpens it, \RF{} grounds metric
induction in checkable labels. Grounding the objective, on whichever axis, is the
common lever.

\noindent\textbf{A reward-free evaluation protocol.} Beyond the method, we
advocate the evaluation \emph{protocol} as the reusable contribution. Reporting
an oracle-reward ceiling, the false-pass rate alongside agreement, a
ranking-versus-calibration split, per-criterion attribution, and
fluency/difficulty stratification gives a far more honest picture of a reward-free
judge than a single accuracy or correlation number. We recommend it for anyone
deploying an LLM judge over agent trajectories, independent of \RF{}.

\noindent\textbf{When induction helps, and threats to validity.} Induction helps
most exactly where generic judges are most dangerous: hard tasks with fluent,
confident agents. It helps least where over-crediting is already rare (easy
retail). Several threats temper the generality of our claims. The 7B judge is
near its binary-agreement ceiling, so the agreement story is genuinely a tie, not
a latent win; the positive class on \taubench{} is small ($\sim$$16\%$), which is
intrinsic to a hard binary success label but inflates the variance of $F_1$ and
$\kappa$; and the \webshop{} calibration loss is real and reported. We also note
that the renderer's faithfulness---surfacing the right observable signals without
leaking the reward---is a load-bearing assumption we audited but that any
deployment must re-audit for its own trajectory format.

\section{Conclusion}\label{sec:conclusion}
We presented \RF{}, a method that \emph{induces} the text of an agent-evaluation
rubric by reflective evolution against ground-truth-labeled trajectories, freezes
it, and applies it as a reward-free metric with one frozen-model call and no
environment access. Our central finding is deliberately honest and contrarian:
on a hard binary success label an induced rubric does \emph{not} significantly
out-agree a generic LLM judge (McNemar $p=0.248$), and it is marginally worse
calibrated in absolute graded score. Where it pays off is \emph{faithfulness}: it
over-credits fluent-but-failed trajectories about half as often
($0.115$ vs.\ $0.173$ false-pass on \taubench{}; three catches, zero reversals),
ranks graded outcomes better (Spearman $0.410$ vs.\ $0.370$), and---because the
artifact is human-readable text---attributes every verdict to the named criterion
responsible. For a reward-free evaluator, where a false pass ships a broken agent
and a false fail merely costs a retry, this is the metric that matters: \RF{} is
\emph{harder to fool than to agree}. By grounding the evaluation objective in
labeled outcomes, induction resists the Goodhart drift toward fluency that an
ungrounded judge silently suffers---the grounded dual of the sibling
meta-objective-collapse result. We release the induction method and a reward-free
evaluation protocol (oracle ceiling, false-pass probe, ranking--calibration
split, per-criterion attribution) as a step toward trustworthy automatic
evaluation of language-model agents.

\noindent\textbf{Limitations.} (i) \emph{Single frozen 7B judge.} All results use
one frozen Qwen2.5-7B model as both agent and judge; whether the faithfulness
advantage transfers to a stronger or larger judge (e.g.\ 30B or a frontier model)
is future work. We expect it to transfer because the induced rubric is
backbone-agnostic \emph{text}---it can be dropped into any judge---but we do not
claim it here. (ii) \emph{Two benchmarks.} We evaluate on \taubench{} and
\webshop{}; broader validation on AgentBench, WebArena, GAIA, and code agents
remains open. (iii) \emph{Small positive class / binary-agreement tie.} The
$\sim$$16\%$ \taubench{} positive rate is intrinsic to a hard binary label and
widens the CIs on $F_1$/$\kappa$; correspondingly the binary-agreement edge over
\Geval{} is directional and not significant, and we do not over-claim it. (iv)
\emph{Calibration.} \RF{} ranks graded outcomes better but is marginally worse in
absolute calibration, so it should be used for agent comparison, not as an
absolute score regressor. (v) \emph{Induction supervision.} Induction needs a
modest pool of labeled trajectories; in the fully label-free regime the method
reduces to its generic seed.

\appendices
\section{Induced Rubrics, Sweep, and Hyperparameters}\label{app:details}

\noindent\textbf{Induction hyperparameters.} GEPA reflective evolution
(reused unchanged~\cite{agrawal2025gepa}) with minibatch $4$ and seed $0$;
budget $200$ rollout calls on \taubench{} (the run used $208$ judging calls and
accepted a pool of $6$ rubrics, best validation agreement $0.769$ at
$20/26$) and $150$ on \webshop{} ($156$ calls, pool of $4$, best validation
$0.750$ at $18/24$). The seed candidate is the generic G-Eval-style rubric, so
accepted rubrics must beat a competent judge. Judge calls are deterministic
(temperature $0$, max $220$ tokens) with a robust-JSON parse and a conservative
$\hat p{=}0$ fallback on any parse/network failure.

\begin{figure}[h]
\centering
\IfFileExists{figures/fig_sweep.pdf}{%
  \includegraphics[width=\columnwidth]{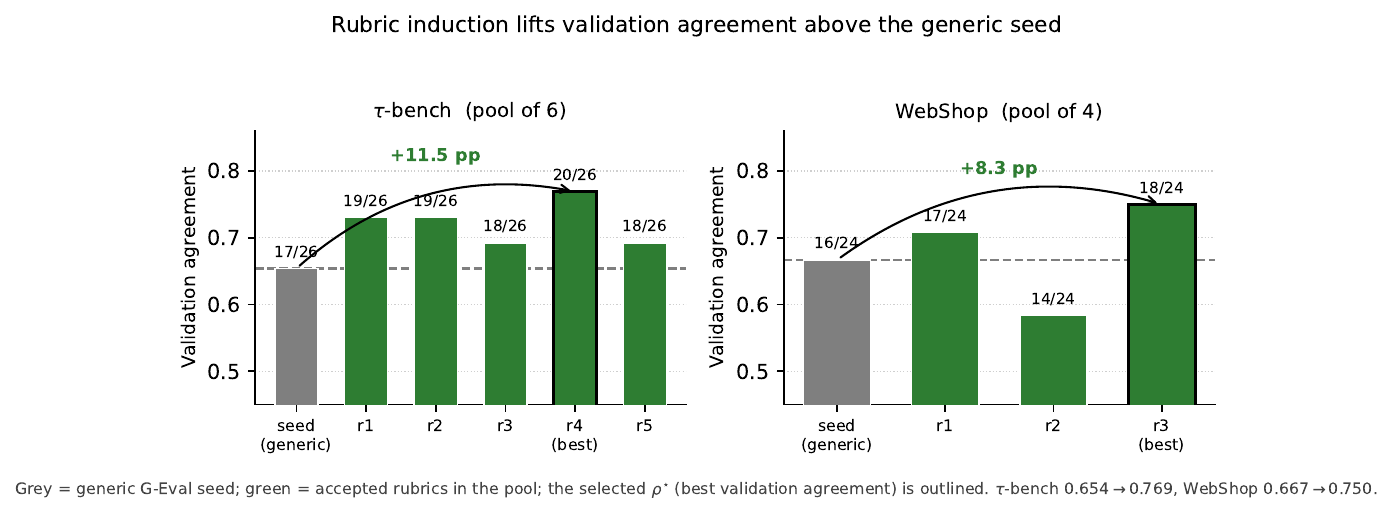}}{%
  \fbox{\parbox[c][3.2cm][c]{0.92\columnwidth}{\centering\texttt{fig\_sweep.pdf}}}}
\caption{Validation-agreement trajectory of rubric induction. Grey is the generic
G-Eval seed ($\rho_0$); green bars are accepted rubrics in the pool; the selected
$\rho^{\star}$ (outlined) lifts validation agreement $0.654\!\to\!0.769$ on
\taubench{} and $0.667\!\to\!0.750$ on \webshop{}. The induced rubric must beat a
competent seed, not a degenerate one.}
\label{fig:sweep}
\end{figure}

\noindent\textbf{Full validation curves.} \Cref{tab:sweep} lists every rubric in
the accepted pool with its validation agreement, and \cref{fig:sweep} plots the
same trajectory. \taubench{}: seed $17/26$, then accepted children at
$19, 19, 18, 20, 18$ ($/26$), best $20/26 = 0.769$. \webshop{}: seed $16/24$,
then $17, 14, 18$ ($/24$), best $18/24 = 0.750$; the corresponding
graded-test ranking-vs-calibration summary is reported in \cref{tab:webshop}.
In both domains every accepted
child except one sits at or above the generic seed, and the selected
$\rho^{\star}$ improves the seed by $+11.5$ and $+8.3$ validation points
respectively---confirming that induction beats a competent starting rubric rather
than a strawman, while the modest absolute gains are consistent with the frozen
7B judge operating near its agreement ceiling.

\begin{table}[h]
\centering
\caption{Induction sweep: validation agreement of every rubric in the accepted
pool (counts out of the validation split; $26$ for \taubench{}, $24$ for
\webshop{}). $\rho_0$ is the generic G-Eval seed; $\rho^{\star}$ (the selected,
frozen rubric) is in \textbf{bold}. Children are listed in acceptance order.}
\label{tab:sweep}
\footnotesize
\renewcommand{\arraystretch}{1.18}
\setlength{\tabcolsep}{4pt}
\begin{tabular}{@{}l ccccccc@{}}
\toprule
Benchmark & $\rho_0$ (seed) & $\rho_1$ & $\rho_2$ & $\rho_3$ & $\rho_4$ & $\rho_5$ & best \\
\midrule
\taubench{} ($/26$) & $17$ & $19$ & $19$ & $18$ & $\mathbf{20}$ & $18$ & $\mathbf{0.769}$ \\
\webshop{} ($/24$) & $16$ & $17$ & $14$ & $\mathbf{18}$ & --- & --- & $\mathbf{0.750}$ \\
\bottomrule
\end{tabular}
\end{table}

\begin{table}[h]
\centering
\caption{\webshop{} graded-test summary ($n=56$). \RF{} wins ranking; \Geval{}
wins absolute calibration (paired-bootstrap $\Delta|{\rm err}|=-0.048$, $95\%$ CI
$[-0.083,-0.018]$, $p=2\!\times\!10^{-4}$).}
\label{tab:webshop}
\footnotesize
\renewcommand{\arraystretch}{1.2}
\begin{tabular}{@{}lccc@{}}
\toprule
Judge & Spearman $\uparrow$ & Kendall $\uparrow$ & mean $|\hat s-\rstar|$ $\downarrow$ \\
\midrule
\textbf{\RF{}} & $\mathbf{0.410}$ & $\mathbf{0.347}$ & $0.336$ \\
\Geval{} & $0.370$ & $0.314$ & $\mathbf{0.288}$ \\
\bottomrule
\end{tabular}
\end{table}

\noindent\textbf{The frozen \taubench{} rubric.}\label{app:rubric} The induced
rubric applied at test time (verbatim, lightly truncated) is:
\begin{quote}\footnotesize\itshape
1. Successful Authentication: the agent must successfully authenticate the user
using at least one valid method; any errors or repeated failed attempts indicate
failed authentication.
2. Order Verification: the agent must retrieve and verify the user's order
details; any ``order not found'' or repeated lookups indicate failure.
3. Request Fulfillment: the agent must correctly fulfill the request (cancel,
modify, return, exchange) under policy, with explicit user confirmation; any
unfulfilled request indicates failure.
4. Tool Usage: the agent must call the appropriate tools without errors; tool-call
errors reduce the score.
5. User Communication: the agent must maintain clear, relevant communication
throughout the interaction.
\end{quote}

\noindent\textbf{The frozen \webshop{} rubric (summary).} Four criteria:
(1)~the search query broadly aligns with the instruction's key attributes;
(2)~the purchased product's attributes (color, size, style) match the request,
allowing only explicitly-justified minor variances; (3)~the purchase respects the
price limit; (4)~the agent completes \texttt{Buy Now} with final options matching
its selections. Failure is indicated if any criterion is unmet.

\noindent\textbf{Reproducibility.} All verdicts are deterministic given the
frozen backbone, the frozen rubric, and the fixed by-task split (seed $0$). The
agreement table, McNemar tests, bootstrap CIs, the over-crediting probe, the
leave-one-out ablation, the calibration split, and the stratification are
computed from the persisted per-item predictions on the held-out test split.

\bibliographystyle{IEEEtran}
\bibliography{references,references_extra}

\end{document}